\documentclass[10pt,twocolumn,letterpaper]{article}

\usepackage{cvpr}              
\usepackage{bm}
\usepackage[dvipsnames]{xcolor}

\definecolor{cvprblue}{rgb}{0.21,0.49,0.74}
\usepackage[pagebackref,breaklinks,colorlinks,citecolor=cvprblue]{hyperref}

\def\paperID{*****} 
\def\confName{CVPR}
\def\confYear{2024}

\title{A Two-Stage Masked Autoencoder Based Network for Indoor Depth Completion}

\author{Kailai Sun $^{1,2}$ \hspace{4em} Zhou Yang $^{1}$ \hspace{4em} Qianchuan Zhao $^{1}$ \hspace{4em}\\
{$^{1}$ Tsinghua University} \hspace{2em} {$^{2}$ National University of Singapore}\\
{\tt\small skl23@nus.edu.sg; yangzhou9$\_$7@163.com; zhaoqc@tsinghua.edu.cn}}
\vspace{-3cm}

\begin{document}
\vspace{-3em}
\maketitle

\begin{abstract}
\vspace{-1em}
Depth images have a wide range of applications, such as 3D reconstruction, autonomous driving, augmented reality, robot navigation, and scene understanding. Commodity-grade depth cameras are hard to sense depth for bright, glossy, transparent, and distant surfaces. Although existing depth completion methods have achieved remarkable progress, their performance is limited when applied to complex indoor scenarios. To address these problems, we propose a two-step Transformer-based network for indoor depth completion. Unlike existing depth completion approaches, we adopt a self-supervision pre-training encoder based on the masked autoencoder to learn an effective latent representation for the missing depth value; then we propose a decoder based on a token fusion mechanism to complete (i.e., reconstruct) the full depth from the jointly RGB and incomplete depth image. Compared to the existing methods, our proposed network, achieves the state-of-the-art performance on the Matterport3D dataset. In addition, to validate the importance of the depth completion task, we apply our methods to indoor 3D reconstruction. The code, dataset, and demo are available at https://github.com/kailaisun/Indoor-Depth-Completion.
\vspace{-0.8em}
\end{abstract}    
\vspace{-1em}
\section{Introduction}
\vspace{-0.8em}
\label{sec:intro}

An investigation by the Environmental Protection Agency (EPA) has found that over 75\% of the global population resides in urban areas and spends approximately 90\% of their time indoors in buildings \cite{EPA}. With the urban population and the proliferation of buildings, there is an increasing demand for the spatial layout of indoor environments and the information about objects within them \cite{human90}. In order to meet the demands of human activities, indoor spatial information representation is of paramount importance. 

Indoor 3D reconstruction \cite{indoor3drecon}, which aims to create a digital spatial information representation of the interior of a building in three dimensions, has gained increasing attention recently. Indoor 3D reconstruction has wide-ranging implications and significance in architecture \cite{Brilakis}, virtual reality, navigation guidance, and robotics. Indoor 3D reconstruction typically uses sensors (e.g., RGB-D camera \cite{RGBD}, laser scanning \cite{laser_scanning}, IMUs \cite{IMUs}) to collect indoor scene data and develop artificial intelligence (AI) methods to effectively capture the geometric details, spatial layout, and semantic understanding of the indoor scene.



\begin{figure*}[!t]
\centering
\includegraphics[width=7.2in]{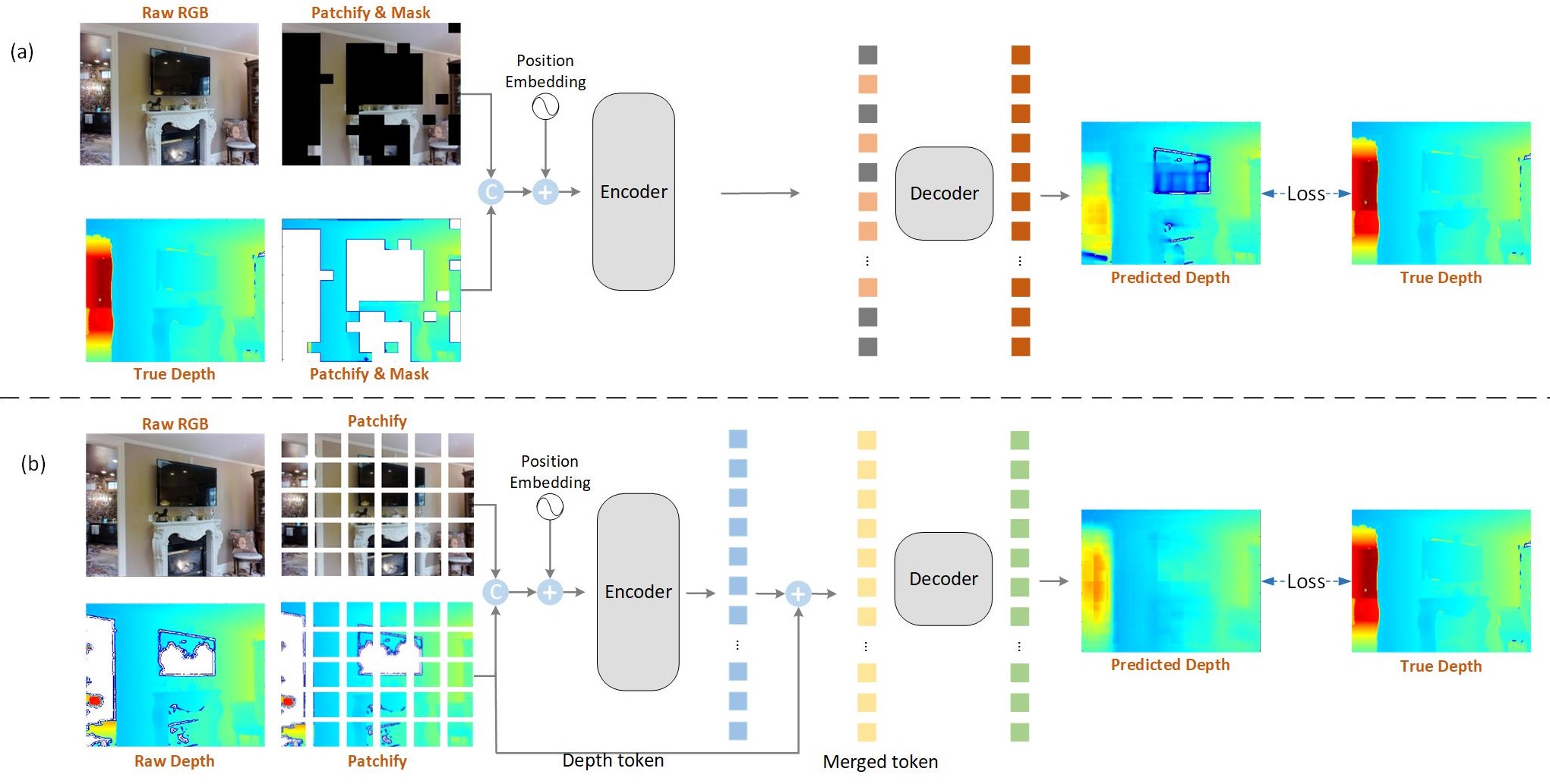}
\vspace{-1em}
\caption{Pipelines of depth completion.}
\vspace{-1.5em}
\label{method}
\end{figure*}

In indoor 3D reconstruction, depth completion is an important task. The depth completion task focuses on using part of the depth data measured in the real scene to obtain more dense and complete depth data. While existing depth completion studies have made notable advancements, their performance is limited when applied to complex indoor scenarios. Traditional methods \cite{tomasi1998bilateral,silberman2012indoor,1013021890.nh,kopf2007joint,ku2018defense} often need many hand-tuning hyper-parameters, and they struggle to fill high-quality depth holes and complete complex depth structures effectively. Existing deep learning-based methods in depth completion can be divided into two categories: only taking the depth map as input \cite{cheng2019learning,huang2019hms,cao2021self,lu2020depth}, and combining the depth map with the RGB image as input. The latter often performs better with more scene information. Early methods in this category typically directly input the depth map and RGB image as four-channel data into the model \cite{ma2018sparse, ma2019self, qu2020depth, long2021depth}. To improve them, recent data fusion methods employ two encoders to process the depth map and RGB image separately. The dual encoders extract multiscale features while performing data fusion \cite{teixeira2020aerial, zhang2020multiscale, Huang,gu2021denselidar}.  However, the latest methods often suffer from sensitivity to dynamic environmental lighting, failing to predict accurate depth completion results under demanding lighting conditions. Moreover, While many depth completion algorithms are tailored for sparse depth images such as LiDAR, they are difficult to directly apply to RGBD cameras. Thus, further research on the network design, and adaptation in complex indoor scenarios and dynamic environments is needed.

To address these problems, we notice that masks in computer vision could be applied to simulate the missing data. Mask-based methods, e.g., Masked Autoencoders (MAE) \cite{he2022masked} only apply partial observation to reconstruct the entire image, making the model learn robust features and improving the generalization ability. Inspired by this, we consider the missing depth patches as masks, then train an encoder-decoder model to complete (i.e., reconstruct) full depth images. We propose a Vision Transformer-based two-stage network for indoor depth completion: an MAE-based self-supervision pre-training encoder to learn an effective latent representation from the jointly masked RGB and depth images; a decoder based on token fusion to complete (reconstruct) the full depth from an incomplete depth image. To our knowledge, it is the first work to apply MAE and Vision Transformers (ViT) in an indoor depth completion task. The contributions: \textit{we introduce an MAE-based self-supervision pre-training mechanism to learn a latent representation for the missing depth value, a supervised fine-tuning mechanism for accurate indoor depth completion.  We demonstrate our method on the Matterport3D dataset with SOTA performance. By applying on the indoor 3D reconstruction task, we highlight the importance of the depth completion task.}

\vspace{-0.8em}

\section{Methods}
\vspace{-0.8em}
Like autoencoder-based depth completion approaches, our method has an encoder that maps the measured depth to a latent representation, and a decoder that completes (i.e., reconstructs) the full depth from the latent representation. Unlike existing depth completion approaches, we adopt a mask-based self-supervision pre-training mechanism to extract an effective latent representation from part depth data; we employ a Vision Transformer encoder, and merge depth tokens to achieve the SOTA performance. Our method includes self-supervision pre-training and supervised fine-tuning. Figure~\ref{method} illustrates the idea, introduced next.

\vspace{-1em}
\subsection{MAE-based Self-supervision Pre-training}\label{MAESSL}

\vspace{-0.8em}
\textbf{Input} In Figure~\ref{method}(a), we follow a self-supervision learning pipeline to learn an effective latent representation from partial observation. We divide a raw RGB image and a true depth image (ground truth) into regular non-overlapping patches, simultaneously. Then, we sample a subset of patches and mask the remaining ones.
We sample random patches from a uniform distribution to prevent more masked patches near the image centre \cite{he2022masked}. We mask these patches to simulate the missing depth in the real scene. The masks in the raw RGB image and the true depth image are spatially aligned. Then, we concatenate the unmasked regions of the RGB image and true depth image to a four-channel RGB-D image for further feature extraction. \\
\textbf{Encoder} The remaining RGB-D image patches after masking are fed into the encoder along with position encodings. Since the masking operation reduces the size of the input data, it allows larger encoder modules to be trained with the same computational cost as the original ViT. We adopt a self-attention mechanism and a Vision Transformer to learn global structural information. Our encoder is based on a Vision Transformer but applied only on unmasked patches. Like a standard Vision Transformer, our encoder embeds patches by a linear projection with added position embeddings. To represent the position of each patch, we use absolute position embedding, a learnable parameter. We straightway add this parameter to the image tokens. Then, we process the tokens via many Transformer blocks. \\
\textbf{Decoder} Before the decoder, because our encoder is applied only on unmasked patches, we need to concatenate the mask tokens with the encoder's output. The masks in RGB-D images are encoded as a shared, learned parameter, which indicates the presence of missing patches to be predicted. We add this parameter and position embeddings to obtain mask tokens. We concatenate the mask tokens with the encoder's output. Then, we process the tokens via our decoder, including many Transformer blocks. \\
\textbf{Loss function} The tokens after the decoder are then mapped to the same dimensions as the original image through a fully connected layer. The purpose of the task is to reconstruct the masked image patches, similar to the depth completion task, which aims to reconstruct/complete the depth image.  In the true depth image, partial depth incompleteness is inevitable. When the model needs to reconstruct parts that contain this incomplete depth, we do not wish the missing depth to mislead the model. Unlike MAE, in the loss function, we only consider the parts where the depth value exceeds zero. In particular, we employ the Root Mean Square Error (RMSE) as the loss function in Equation \eqref{eq:rmse}:
\vspace{-0.8em}
\begin{equation}
  RMSE\left(D,D_T\right) = \sqrt{\frac{1}{\lvert O \rvert}\sum_{p \in O}\Vert D\left(p\right) -D_T\left(p\right) \Vert^2},
  \label{eq:rmse}
\end{equation}
where $D$ represents the predicted depth image, $D_T$ represents the true depth image, $p$ represents a pixel in the depth image, and $O$ represents non-zero pixels in $D_T$.


\subsection{Indoor Depth Completion based on Supervised Fine-tuning}
\vspace{-0.8em}
Using this pretrained encoder and decoder directly for depth image completion still has many limitations. The masked RGB image patches have yet to be utilized for depth prediction (i.e., underuse the full RGB image). Prior information from the RGB image has significant guiding implications \cite{9984942}. Besides, the patches are square, while missing depth areas are irregular, resulting in the information loss of RGB-D images. Therefore, to overcome these limitations, we consider our encoder as a pre-training model to extract implicit features and finetune the pretrained model to complete the entire image. Figure~\ref{method}(b) illustrates the whole algorithm pipeline. The algorithm(b) is similar to the algorithm(a); the main differences:\\
\textbf{Input} In the fine-tuning process, following the self-supervised learning method, we feed the masked true depth image into the encoder and reconstruct the true depth image conditional on the masked RGB image. In contrast, we use a supervised learning method to complete the depth image using the raw RGB and depth images. Thus, the input only contains RGB-D image patches without masks.\\
\textbf{Token Fusion} Before the pre-training encoder, the RGB/depth image patches are fused by concatenation. We perform multi-modal data fusion at both the input level and the token level. After the pre-training encoder, the encoder's output and depth tokens are fused. We add a residual connection to connect the raw depth image with the intermediate tokens from the encoder. In this step, the features extracted from the RGB image and the depth image can be fused with the depth image for the second time.\\
\textbf{Loss function} After the decoder, the tokens are mapped to the same size as the raw depth image through a fully connected layer. Different from \eqref{eq:rmse}, we compute the loss in all patches, including non-zero pixels and zero pixels. Through the above improvements, our model can utilize the full RGB image information and carry out two-stage data fusion in the whole process.

\vspace{-0.8em}
\section{Experiments and Discussion}
\vspace{-0.8em}
\subsection{EXPERIMENTAL SETUP}
\vspace{-0.8em}
\textbf{Dataset} We use the famous dataset from work \cite{zhang2018deep}, which is modified on the original MatterPort3D \cite{chang2017matterport3d}.  The original data set takes RGB and depth panoramas at fixed points indoors and then aligns RGB and depth images. We follow the method \cite{zhang2018deep} to generate true depth images. As a result, the dataset includes 104699 images for training and 474 images for testing. The size of each image is $320\times256$. \\
\textbf{Evaluation Indicators} We use the RMSE in the section~\ref{MAESSL}, Mean Error(ME), Structure Similarity Index Measure(SSIM \cite{SSIMmea}), and $\delta_t$ for performance comparison. \\
\textbf{Hyperparameters} The input images are resized to $224\times224$. The channel is 4. We set the mask rate in our pre-training model at 75\%. The encoder comprises 24 layers with 16 attention heads and an embedding dimension of 1024. The decoder comprises 8 layers with 16 attention heads and an embedding dimension of 512. Our pre-training model is trained for 200 epochs while our fine-tuning model is trained for 20 epochs.

\begin{table*}[!h]
  \centering
  \setlength{\tabcolsep}{4pt}
  \caption{Quantitative comparison of our proposed methods against the existing SOTA methods on the Matterport3D dataset. RMSE and MAE are measured in meters.}
 \vspace{-1em}
  \begin{tabular}{p{5cm}p{1.5cm}p{1.5cm}p{1.5cm}p{1.5cm}p{1.5cm}}
    \toprule
    Methods & RMSE$\downarrow$ &ME$\downarrow$ &SSIM$\uparrow$&$\delta_{1.25}$ $\uparrow$&$\delta_{1.25^2} \uparrow$  \\
    \midrule
  Joint Bilateral Filter  &1.978 & 0.774 &0.507 &0.613& 0.689\\
  MRF\cite{MRF}&1.675& 0.618&0.692 &0.651& 0.780\\
  AD\cite{A11}& 1.653& 0.610 &0.696& 0.663 &0.792\\
  FCN& 1.262& 0.517& 0.605&0.681& 0.808\\
  Zhang\cite{zhang2018deep}  &1.316 &0.461 &0.762&0.781 &0.851 \\
  Huang\cite{Huang}  &1.092 &0.342 &$\bm{0.799}$& 0.850& 0.911\\
   Struct-MDC\cite{jeon2022struct}&$1.060$&0.503&0.534&0.656&0.713\\
\hline
  Pre-training  &1.216 &0.675 &0.642 &0.705 &0.800 \\
 Fine-tuning  w/o Pre-training   &$\bm{0.660}$ &0.243 &0.654 &0.794 &0.904 \\
  Fine-tuning  w/ Pre-training  &$\bm{0.690}$ &$\bm{0.206}$ &$\bm{0.765}$&$\bm{0.852}$&$\bm{0.912}$ \\
    \bottomrule
  \end{tabular}
  \label{tab:comple_compare}
\vspace{-1em}
\end{table*}

\begin{figure*}[!ht]
  \centering
\includegraphics[width=0.99\linewidth]{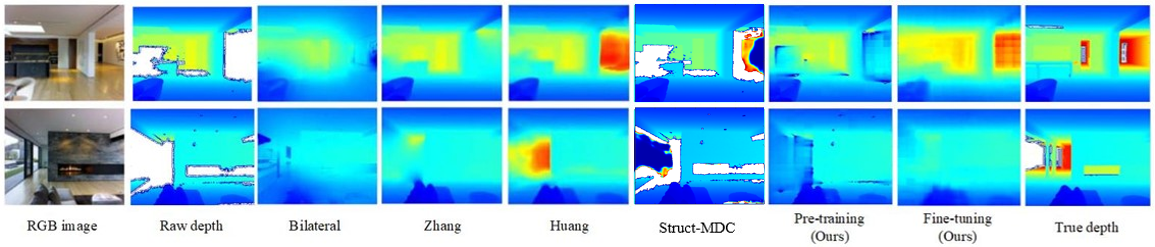}
  \vspace{-1em}
  \caption{Qualitative comparison of our proposed methods against the existing depth completion methods on the Matterport3D dataset.}
  \vspace{-1em}
\label{fig:comple_compare}
\end{figure*}

\vspace{-0.8em}
\subsection{Quantitative and Qualitative comparison}
\vspace{-0.8em}
\begin{table*}[!h]
    \centering
    \caption{3D Reconstruction Errors}
    \vspace{-1em}
    \begin{tabular}{cccccc}
    \toprule
    Methods&Mean (m)$\downarrow$&Median (m)$\downarrow$&Standard Deviation (m)$\downarrow$&Minimum (m)&Maximum (m)$\downarrow$\\
    \midrule
    Depth uncompletion&0.138 &0.053 &0.200 &0.0&1.106\\
    Depth completion&$\bm{0.086}$&0.057 &$\bm{0.101}$ &$\bm{0.0}$&$\bm{1.100}$\\
    \bottomrule
    \end{tabular}
    \label{tab:jieheresult}
    \vspace{-2em}
\end{table*}

In Table~\ref{tab:comple_compare}, we compare our proposed methods with some existing depth completion methods, including traditional methods (e.g., joint bilateral filter, JBF), FCN-based methods (ResNet18), the methods of Zhang ~\cite{zhang2018deep} and ~\cite{Huang}, and Struct-MDC\cite{jeon2022struct}. Compared to the existing methods, our proposed methods, achieve the state-of-the-art (SOTA) performance on the Matterport3D dataset. The result shows our significant improvement in most of the evaluation metrics (e.g., RMSE, ME and $\delta_t$).

The pre-training model is better than the traditional methods (e.g., MRF) and the method of Zhang~\cite{zhang2018deep} on RMSE, and only better than the Joint bilateral filter on ME, and better than Joint bilateral filter and FCN on SSIM. 

Our fine-tuning model achieves a significant improvement. More importantly, our fine-tuning model achieves superior performance on the Matterport3D dataset. In particular, on $\delta_t$ and ME, our fine-tuning model performs best. In Figure ~\ref{fig:comple_compare}, our method learns the structural features in the scene effectively, reconstructing the edge structure instead of only smoothing the missing depth values.

Our fine-tuning model also has limitations. It is slightly inferior to the method of Huang\cite{Huang} on SSIM. SSIM focuses on the overall structure of graphics and the intuitive feeling of human eyes\cite{SSIMmea}. And SSIM can be easily affected by variance and mean. For example, in Figure ~\ref{fig:comple_compare}, the depth images generated by our Transformer models have unsmooth grids, which need to be improved in the future.

\vspace{-1em}
\section{An application:Indoor 3D Reconstruction}
\vspace{-0.5em}

Indoor 3D reconstruction by RGB-D cameras has many errors because of camera limitations (e.g., bright lighting conditions, limited measured range, and depth images often contain holes and noise). We claim that deep completion plays a crucial role in ensuring the accuracy and reliability of indoor 3D reconstruction.  We use a living room scene in the ICL-NUIM dataset \cite{handa:etal:ICRA2014} to confirm.

As for the input, for a fair comparison, we adopt an image fusion strategy in Equation \ref{eq:ronghe} to merge the depth image from our depth completion network with the original image. 
\begin{equation}
  {D_R}\left(p\right) = \left\{
\begin{array}{lll}
D_O\left(p\right)      &      & {D_O\left(p\right) \neq 0},\\
D\left(p\right)     &      & {D_O\left(p\right) = 0},
\end{array} \right.
  \label{eq:ronghe}
\end{equation}
where $D_R$ represents the fused depth image, $D_O$ represents the original depth image, and $D$ represents the depth image, which is outputted by our depth completion network.

As for evaluation indicators, we followed the evaluation method recommended by the ICL-NUIM dataset, which uses the point cloud distance measurement in the open-source software (i.e., CloudCompare \cite{girardeau2016cloudcompare}) to quantify the reconstruction quality of 3D models.


We choose ORB SLAM3 \cite{rgbslam3} to estimate pose estimation and then use TSDF volume and marching cubes algorithm\cite{marching}  to get 3D reconstruction models. Since the estimated trajectories exhibit minimal disparity, we conduct separate experiments to compare the 3D reconstruction results using the completed depth image $D_R$  or the original, incomplete depth image $D_O$. The errors are shown in Table \ref{tab:jieheresult}. It can be observed that the depth completion method leads to a more minor mean error, standard deviation, and maximum error. Only the median error slightly increases. Overall, ORB SLAM3 with completed depth images performs better than incompleted depth images.

\vspace{-0.8em}
\section{Conclusion}
\vspace{-0.5em}
Although depth cameras can sense the distance information between objects, many depth values are often lost because of the bright, glossy, transparent, and distant surfaces. In this paper, we propose a two-stage Transformer-based network to complete an indoor depth image given a single RGB-D image. Our proposed network achieves the SOTA performance on the Matterport3D dataset. An indoor 3D reconstruction task application highlights the depth completion's importance. We believe our method can be widely applied in other areas, including 3D understanding, building renovation or construction, etc.
{
    \small
    \bibliographystyle{ieeenat_fullname}
    \bibliography{main}
}

\end{document}